\documentclass[conference]{IEEEtran}
\IEEEoverridecommandlockouts
\usepackage{cite}
\usepackage{amsmath,amssymb,amsfonts}
\usepackage{algorithmic}
\usepackage{graphicx}
\usepackage{textcomp}
\usepackage{xcolor}
\usepackage{dsfont}
\usepackage{comment}
\usepackage{upgreek}
\def\BibTeX{{\rm B\kern-.05em{\sc i\kern-.025em b}\kern-.08em
    T\kern-.1667em\lower.7ex\hbox{E}\kern-.125emX}}
    
\usepackage{fancyhdr}
\begin{document}

\title{AffRankNet+: Ranking Affect Using Privileged Information\\
\thanks{This work has been supported by the European Union's Horizon 2020 research and innovation programme from the TAMED project (Grant Agreement No. 101003397).}
}

\author{\IEEEauthorblockN{Konstantinos Makantasis}
	\IEEEauthorblockA{\textit{Institute of Digital Games} \\
		\textit{University of Malta}\\
		Msida, Malta \\
		konstantinos.makantasis@um.edu.mt}
}

\maketitle
\thispagestyle{fancy}

\begin{abstract}
Many of the affect modelling tasks present an asymmetric distribution of information between training and test time; additional information is given about the training data, which is not available at test time. Learning under this setting is called Learning Under Privileged Information (LUPI). At the same time, due to the ordinal nature of affect annotations, formulating affect modelling tasks as supervised learning ranking problems is gaining ground within the Affective Computing research community. Motivated by the two facts above, in this study, we introduce a ranking model that treats additional information about the training data as privileged information to accurately rank affect states. Our ranking model extends the well-known RankNet model to the LUPI paradigm, hence its name AffRankNet+. To the best of our knowledge, it is the first time that a ranking model based on neural networks exploits privileged information. We evaluate the performance of the proposed model on the public available Afew-VA dataset and compare it against the RankNet model, which does not use privileged information. Experimental evaluation indicates that the AffRankNet+ model can yield significantly better performance.
\end{abstract}

\begin{IEEEkeywords}
Ranking affect, preference function, privileged information, knowledge distillation, RankNet
\end{IEEEkeywords}

\section{Introduction}
\label{sec:intro}
One of the most popular ways for annotating affect is based on rating systems, such as simple Likert scales \cite{likert1932technique}, self-assessment manikins \cite{morris1995observations}, and rating scales of the discrete states in the Geneva emotion wheel \cite{scherer2005emotions}. The common characteristic of all the above rating systems is that they provide \textit{ordinal} and not \textit{nominal} information about the affect states. In addition, several psychometric studies show that ratings of affect do not follow an absolute and consistent scale \cite{ovadia2004ratings,metallinou2013annotation}. Therefore, learning to predict nominal values of affect yields inconsistent models of questionable quality and use. On the contrary, treating ratings as ordinal values yields less biased datasets and, thus, more reliable models of affect \cite{martinez2014don}. For the reasons above, formulating affect modelling tasks as supervised learning ranking problems is gaining ground within the Affective Computing research community.

The supervised learning problem of ranking consists of using labelled information to derive accurate ranking prediction functions. Most of the algorithms that try to address that problem are using information that comes solely from labelled pairs of data points and transform the ranking problem into a classification one \cite{joachims2002optimizing,freund2003efficient,burges2010ranknet}. The label of a pair $(x_i, x_i')$ of data points is 1, -1, or 0 if $x_i$ is ranked higher than, lower than, or equal to $x_i'$, respectively. Hence, these algorithms can be used even when only the global ordering of data points is provided without the need for preference scores or other kinds of information.

In many real-world applications, however, there is an asymmetric distribution of information between training and test time; that is, additional information is given about the training data, which is not available at test time. Consider, for example, user ratings for different movies, self-assessment manikin scale for affect annotation, or the number of likes and dislikes associated with advertisements. Although this additional information, which can be implicitly seen as preference scores, is very valuable, it is disregarded by algorithms that use solely labelled pairs of data points. 

In this study, we propose a supervised learning ranking model of affect. Besides the information that comes from labelled pairs of data points, our model also exploits additional information that directly or indirectly is associated with preference scores, that is ordinal values of affect states. Since this additional information can only be available during the training phase of the model and not at test time, we treat it as \textit{privileged information} and follow the learning paradigm of Learning Under Privileged Information (LUPI) proposed by Vapnik and Vashist \cite{vapnik2009new}. Our model of affect is based on Neural Networks (NN) and extends the well-known RankNet \cite{burges2005learning} to the LUPI paradigm, hence its name \textit{AffRankNet+}. To the best of our knowledge, this is the first time that privileged information is incorporated into NN for addressing supervised learning ranking problems and the first time that the LUPI paradigm is used for affect modelling. Experimental validation of AffRankNet+ on the large scale publicly available Afew-VA dataset \cite{kossaifi2017afew} indicates that privileged information \textit{significantly} improves the ranking performance of affect models.

\section{Related Work}
This section surveys literature on supervised learning ranking models, and affect modelling based on ranking/preference learning approaches.

\subsection{Supervised Learning Ranking Models}
The supervised learning problem of ranking, based on labelled pairs of data points, has been widely studied. Below we present some landmark works focusing on this problem. 

RankSVM proposed in \cite{joachims2002optimizing} was one of the first approaches focusing on this problem. The authors use Support Vector Machines (SVM) to compute a preference function. 
In \cite{kuo2014large, lee2014large} the authors reduce the number of RankSVM variables from quadratic to linear with respect to the number of training instances in order to significantly reduce the training time and make RankSVM suitable for large-scale problems.

RankBoost \cite{freund2003efficient, rudin2009margin, connamacher2020rankboost} is another well-known ranking algorithm. RankBoost creates and aggregates a set of ranking functions in an iterative fashion to build an effective ranking procedure. Using solely information that comes from labelled pairs of data points, RankBoost estimates a preference function that can map single points to real-valued preference scores. 

The authors in \cite{burges2005learning} approach the ranking problem by proposing a probabilistic cost function for training machine learning models. In their study, they utilize NN, and thus they call their approach RankNet. The idea, however, of employing a probabilistic cost function has equally well been applied to ranking algorithms that adopt different learning machines, such as Boosted Trees \cite{burges2010ranknet}. Similarly to the approaches presented above, RankNet is trained on labelled pairs of data points. After training, it can evaluate single points and produce preference scores for each one of them. DeepRank \cite{pang2017deeprank}, which targets information retrieval tasks, is also based on NN. However, it differs from RankNet, since it identifies and exploits local preference relations between the data points to induce the global ranking. In \cite{rahangdale2019deep} the authors introduce $l_1$ regularization to a NN-based ranking model, to enforce sparsity and avoid overfitting. Since the above mentioned approaches are based on NN, they can straightforward exploit the recent advances in deep learning \cite{song2014adapting, parthasarathy2017ranking} and tensor-based learning \cite{makantasis2018tensor, makantasis2019common, makantasis2021rank}. However, none of these follows the LUPI paradigm to exploit additional/privileged information about the training data that might be available. In other words, they follow the typical supervised learning setting by transforming the ranking problem to a classification one.

Selecting a preference function using the methods presented above is based solely on the order of the data points. Even if additional information is available, such as preference scores associated with the points, this information is entirely disregarded. In this study, we argue that exploiting additional information associated with preference scores can produce more accurate ranking algorithms. We assume that the additional information is available only during the training phase of the model and not at test time. This assumption is critical to impose no restrictions related to capturing additional information during the real-world deployment of the model. To enable the AffRankNet+ model to exploit additional information during training efficiently, we follow the LUPI paradigm \cite{vapnik2009new, vapnik2015learning}, which is closely related to knowledge distillation proposed in \cite{hinton2015distilling}. Theoretical results \cite{lopez2015unifying, pechyony2010theory} show that following the LUPI paradigm reduces the sample complexity of the learning algorithm, which implies that LUPI models learn faster, and at the same time, they are very efficient for small sample setting problems, i.e. problems where the number of annotated samples is limited.


\subsection{Ranking-based Affect Modelling}
Based on psychological theories and evidence from multiple disciplines, such as neuroscience and artificial intelligence, Yannakakis et al. \cite{yannakakis2018ordinal} draw the theoretical reasons to favour ordinal labels for representing and annotating affective states. They also suggest ranking/preference learning as the appropriate approach for building reliable and valid affect models. Due to the ordinal nature of emotions, several studies approach affect modelling using ranking or preference machine learning algorithms.  

In \cite{yang2010ranking} the authors represent the emotion elicited by a music song as a point into a two-dimensional Cartesian space with valence and arousal as dimensions. The coordinates of a song are determined relatively, using a modification of the ListNet \cite{xia2008listwise} algorithm, with respect to other songs’ emotions. The study in \cite{fan2017ranking} also focuses on music emotion recognition. The authors first collect a dataset and annotate it using ordinal labels. Then, they propose a modification of RankSVM, called smoothed RankSVM, for deriving emotion recognition models.

Similarly, the study in \cite{zhou2018relevant} proposes a ranking algorithm to identify the emotions that are more intensely associated with a given text. By exploiting the ordinal nature of emotions, their proposed approach outperforms multi-label classification methods. The authors in \cite{liang2018multimodal} propose a multimodal ranking algorithm for emotion recognition. Their algorithm is based on the emotion intensity gradient; that is, the relative emotion intensity change between two or more different inputs. The authors in \cite{soleymani2008affective} exploit a ranking algorithm to predict spectators’ felt emotions for a given movie scene. They use both physiology and audio-visual features to build and evaluate their models of affect. In \cite{makantasis2021pixels} audio-visual information from gameplay videos is fed to a deep learning RankNet model to estimate the intensity of emotions felt by gamers while they were playing a game. Finally, due to the theoretical and experimental evidence that ordinal data processing yields more reliable, valid and general models of affect, the authors in \cite{camilleri2019pyplt} present the open-source Python Preference Learning Toolbox (PyPLT) to enable the extensive use of ordinal data processing and ranking algorithms.

\begin{figure*}[!tb]
	\begin{minipage}{0.33\linewidth}
		\centering
		\centerline{\fbox{\includegraphics[width=0.95\linewidth]{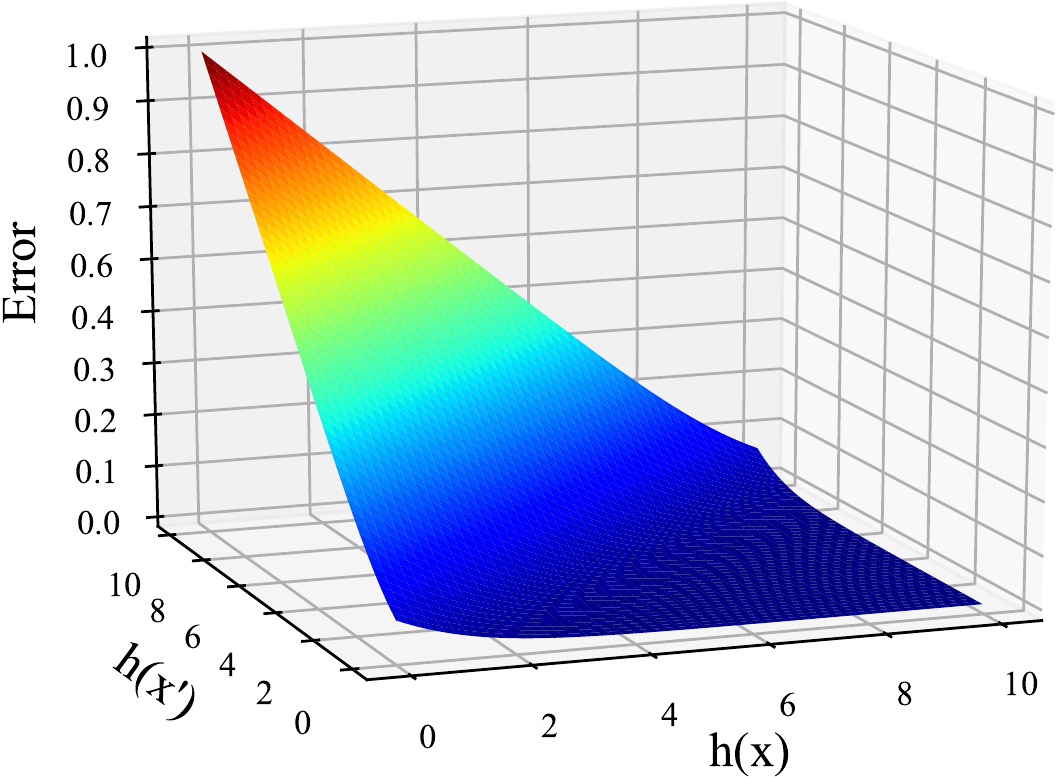}}}
	\end{minipage}
	\begin{minipage}{0.33\linewidth}
		\centering
		\centerline{\fbox{\includegraphics[width=0.95\linewidth]{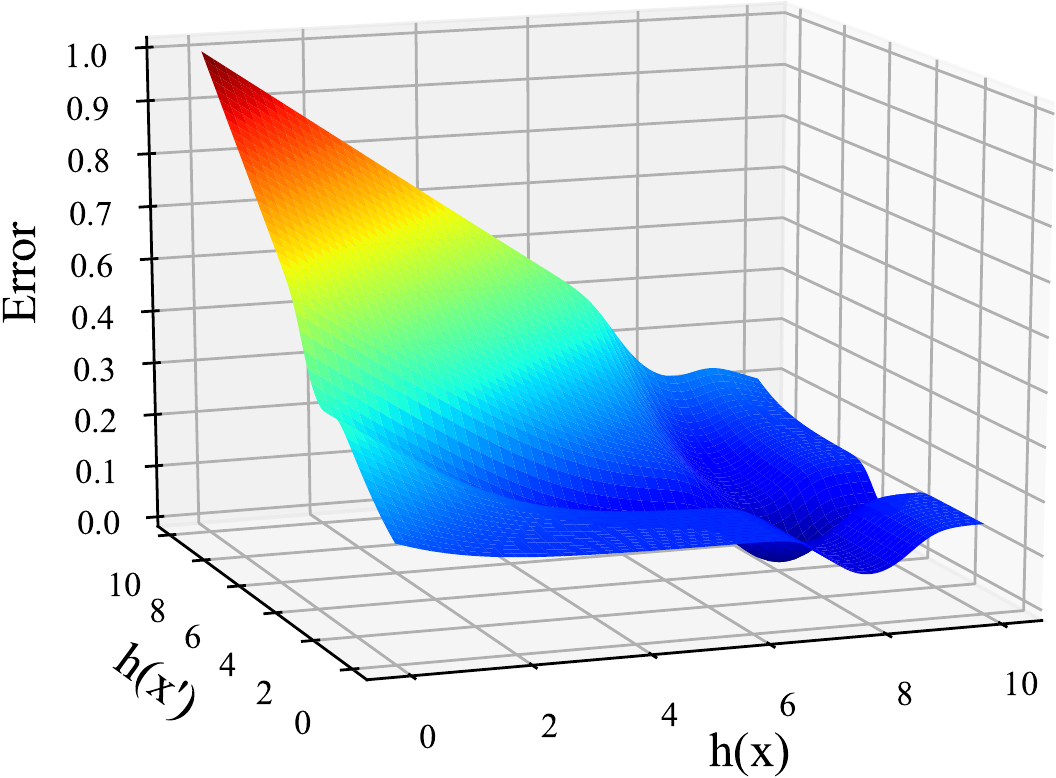}}}
	\end{minipage}
	\begin{minipage}{0.33\linewidth}
		\centering
		\centerline{\fbox{\includegraphics[width=0.95\linewidth]{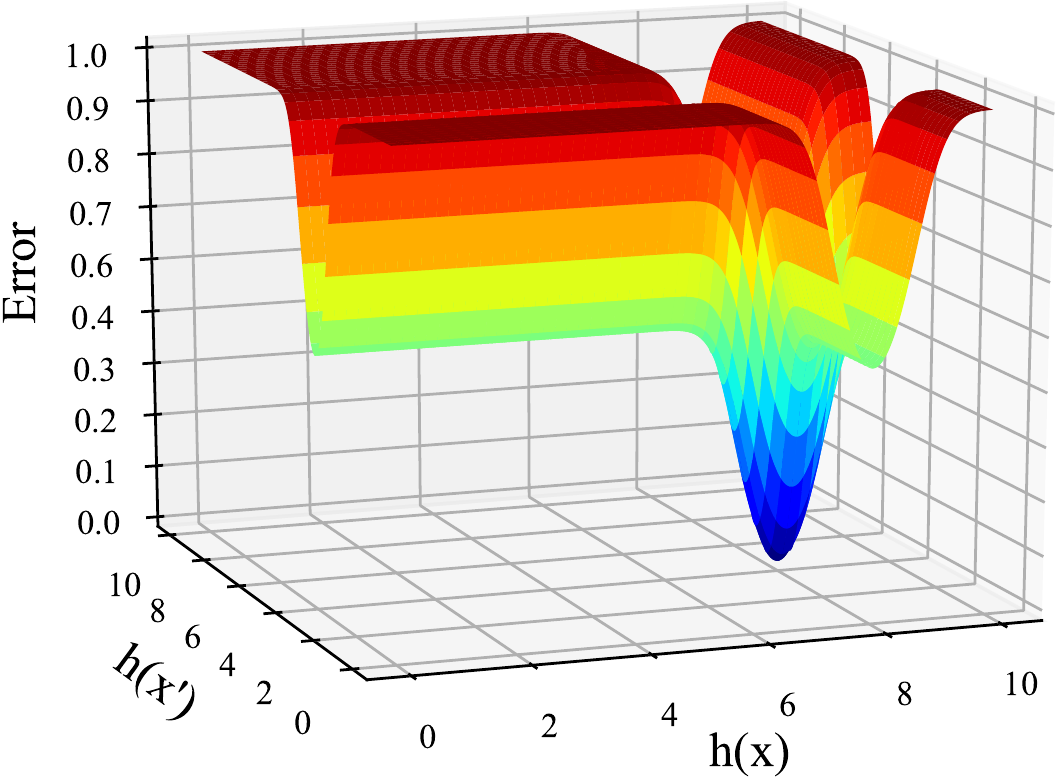}}}
	\end{minipage}
	\caption{Error surfaces normalized to [0,1] for the losses in (\ref{eq:empirical_error}) (left) and (\ref{eq:empirical_error_lupi}) (middle) for a pair of points $(x, x')$ when $f(x, x')=1$, $g(z)=8$, $g(z')=4$, $\lambda=0.5$ and $\tau=1$. The diagram on the right presents the error due to the terms that correspond to the privileged information in (\ref{eq:empirical_error_lupi}).}
	\label{fig:costs}
\end{figure*}

\subsection{Our Contribution}
The contribution of this study is four-fold. First, to the best of our knowledge, we propose for the first time an NN-based supervised learning algorithm that focuses on the problem of ranking and exploits privileged information associated with preference scores. Second, since our approach utilizes NN, it can take full advantage of the recent advances in deep learning and tensor-based NN, such as automatic feature extraction and information processing in high dimension spaces. The above implies that our model can be straightforwardly applied to data points that are represented as feature vectors, but also to data points that lie in tensor spaces such as images, videos, multi-model data (e.g. audiovisual signals) as well as to spatiotemporally evolving sensor network data \cite{makantasis2020space}. Third, by exploiting additional information only during the training phase, the potential applications of our model are not restricted by the requirement of capturing additional information at deployment time. Fourth, we evaluate the proposed model on a large scale publicly available affect dataset; the evaluation results indicate that exploitation of privileged information significantly improves ranking results.

\section{Problem Formulation}
In this section, we first present the ranking problem when information comes solely from labelled pairs of data points. Then, we present its extension to follow the LUPI paradigm assuming privileged information regarding preference scores is available only during the training phase of the model (and not during test time). For simplicity, we formulate the problem of ranking based on labelled pairs of data points as a binary classification problem. However, the formulation can be straightforwardly modified to treat this problem as a three-class classification task to consider pairs of points that are non-comparable or equally preferred

\subsection{The Ranking Problem}
Let us denote by $\mathcal X$ the input space i.e. the feature space of data points, by $f: \mathcal X \times \mathcal X \rightarrow \{0,1\}$ a target labeling  function, and by "$\succ$" and "$\preceq$" preference relations; $x_i \succ$ $x_j$ means $x_i$ is ranked higher than $x_j$ and thus $f(x_i,x_j)=1$. Similarly, $x_i \preceq$ $x_j$ means that $x_i$ is ranked lower than or equal to $x_j$ and thus $f(x_i,x_j)=0$. Given a set of labelled points
\begin{equation}
	\label{eq:sample}
	S = \{(x_i, x_i', t_i)\}_{i=1}^m,
\end{equation}
where $t_i = f(x_i, x_i')$, and a class $\mathcal H$ of preference functions mapping $\mathcal X$ to $\mathbb R$, the Empirical Risk Minimization principle (ERM) \cite{vapnik1999overview} suggests to select a preference function $h^* \in \mathcal H$ that minimizes the empirical error (error over the training set $S$), i.e.
\begin{equation}
	h^* \in \arg \min_{h \in \mathcal H} \hat{R}_S(h),
\end{equation}   
where $\hat{R}_S(h)$ stands for the empirical error of the preference function $h$ and can be quantified by the Binary Cross Entropy (BCE) loss function
\begin{equation}
	\label{eq:empirical_error}
	\hat{R}_S(h) = -\frac{1}{m} \sum_{i=1}^m t_i \log(p_i) + (1-t_i) \log(1-p_i),
\end{equation}
where $p_i=\sigma(h(x_i)-h(x_i'))$, and $\sigma(x)=1/(1+\exp(-x))$ is the sigmoid function. 

At this point, we should mention that the class of functions $\mathcal{H}$ contains all the functions that a given machine learning model can compute. Consider, for example, a neural network with a given architecture. Then every function that the above neural network can compute for different values for its weights belongs to $\mathcal{H}$.

\subsection{The Problem of Ranking Using Privileged Information} 
LUPI is based on the availability of additional information, called \textit{privileged information}, that can be used only during the training phase of a learning model. According to LUPI, exploitation of privileged information during training makes a learning model learn better and faster \cite{vapnik2009new, vapnik2015learning}. This information, however, is not available at test time.

Let us denote as $\mathcal Z$ the space of privileged information. Then, the set of labelled points in (\ref{eq:sample}) is enhanced by the presence of privileged information as
\begin{equation}
	\label{eq:sample_lupi}
	S_{LUPI} = \{(x_i, x_i', z_i, z_i', t_i)\}_{i=1}^m,
\end{equation}
where $z_i, z_i' \in \mathcal Z$, and in general $\mathcal X \neq \mathcal Z$. 

Considering especially the problem of ranking, $z_i$'s should correspond to a representation of information that can be used to estimate preferences scores for $x_i$'s (for example, the output of a learning model that has been trained on $z_i$'s to predict preference scores), or to a direct representation of those preference scores. As far as the latter case is concerned, having available $z_i$'s, which are a direct representation of preference scores, is prevalent for many real-world ranking applications; consider, for example, affect ratings from ordinal annotation tools be directly used as preference scores. 

In the following, we unify the two cases of privileged information mentioned above by considering a function $g: \mathcal Z \rightarrow \mathbb R$ that transforms $z_i$'s to preference scores. In the second case where $z_i$'s are a direct representation of preference scores, $g$ is the identity function, i.e. $g(z_i) = z_i$. The function $g$ in LUPI and knowledge distillation parlance is called ``teacher".

For exploiting privileged information we modify the empirical error in (\ref{eq:empirical_error})) as follows
\begin{equation}
	\label{eq:empirical_error_lupi}
	\begin{split}
		&\hat{R}_{S_{L}}(h) = -\frac{\lambda}{m} \sum_{i=1}^m t_i \log(p_i) + (1-t_i) \log(1-p_i) + \\ &(1-\lambda)(\phi(\frac{(h(x_i)-g(z_i))^2}{\tau}) + \phi(\frac{(h(x_i')-g(z_i'))^2}{\tau})),
	\end{split} 
\end{equation}
where function $\phi$ is the hyperbolic tangent function, i.e., $\phi(x)=\tanh(x)$, that bounds the additional error terms to $[0,1)$, $\lambda \in [0,1]$  is balancing the error terms, and $\tau>0$ is a temperature parameter that quantifies the degree to which the values of the preference scores can be trusted. Fig. \ref{fig:costs} presents the error surfaces normalized to $[0,1]$ for equations (\ref{eq:empirical_error}) and (\ref{eq:empirical_error_lupi}), when $\lambda=0.5$ and $\tau=1.0$ and the preference scores for the two data points are $8$ and $4$, respectively. The same figure also presents the error added to the cost due to the two additional terms in (\ref{eq:empirical_error_lupi}) which corresponds to the privileged information that comes in the form of preference scores. While the loss in (\ref{eq:empirical_error}) considers as best solution the one that maximizes the difference $h(x)-h(x')$, the loss in (\ref{eq:empirical_error_lupi}) selects the preference function $h$ that at the same time reduces the BCE loss and matches, as match as possible, the preference scores provided by the privileged information.

Based on the discussion above, given a labelled set of training points in the form of equation (\ref{eq:sample_lupi}) and a set of preference functions $\mathcal H$, the main objective of this study is to select a function $h^* \in \mathcal H$ that minimizes the loss in (\ref{eq:empirical_error_lupi}).

A natural question that arises is why not use a typical regression model for estimating the preference function $h$ using as target the preference scores provided by the privileged information. Minimizing the sum of BCE and the last two terms of equation (\ref{eq:empirical_error_lupi}) determines simultaneously the distance of a labelled pair of points from the classification decision boundary, and the degree to which the preference function $h(\cdot)$ matches the preference scores coming from privileged information. In ranking problems, the preference scores are usually subjectively biased; different users follow a different internal/personal preference function for providing their ratings. Parameters $\lambda$ and $\tau$ in equation (\ref{eq:empirical_error_lupi}) determine the degree to which the learning model should trust the provided preference scores. Doing the same thing using a typical regression model is not possible. In addition, the loss in (\ref{eq:empirical_error_lupi}) is not just balancing classification and regression losses; BCE and mean squared error. The employment of a general function $g:\mathcal Z \rightarrow \mathbb R$ that transforms privileged information to preference scores indicates that our model goes beyond the typical supervised learning setting to benefit from the properties (robust and fast training) of the LUPI paradigm.  

\section{The AffRankNet+ Model}
Our proposed model is based on and extends the RankNet model \cite{burges2005learning}. Like RankNet, it is an NN-based learning model that is trained on labelled pairs of data points. At the same time, however, and unlike RankNet, it exploits privileged information related to preference scores during its training phase to learn faster and in a more robust way.

Specifically, our proposed model is a two-stream neural network. The architectures of the neural networks corresponding to the two streams are identical, and their weights are tied. Therefore, the two streams produce the same outputs, given that their inputs are the same. Let us denote by $\hat{h}(x)$ the output of each stream when $x$ is given as input. The model receives as input a labelled pair of data points, that is $(x_i, x_i', g(z_i), g(z_i'), t_i)$. The first data point $x_i$ is fed as input to the first stream, which outputs $\hat{h}(x_i)$. Similarly, the second point $x_i'$ is fed as input to the second stream, which outputs $\hat{h}(x_i')$. Having the outputs of the two streams, the model predicts a label $\hat{t}_i$ for the sample $(x_i, x_i')$ as follows:
\begin{equation}
	\hat{t}_i = 
	\begin{cases}
		\:\:\:1  \:\:\text{ if } \sigma(\hat{h}(x_i) - \hat{h}(x_i')) > 0.5\\
		\:\:\:0  \:\:\text{ otherwise}
	\end{cases},
\end{equation}
where $\sigma$ is the sigmoid function. The output $\hat{t}_i$ replaces $t$ in (\ref{eq:empirical_error_lupi}) and is used to estimate the missranking error. For computing the second and the third terms in equation (\ref{eq:empirical_error_lupi}), we replace $h(x_i)$ and $h(x_i')$ with $\hat{h}(x_i)$ and $\hat{h}(x_i')$ respectively. 

Given that we have in our disposal a training set of labelled points in the form of equation (\ref{eq:sample_lupi}) and a function $g:\mathcal Z \rightarrow \mathbb R$ that transforms $z_i$'s to preference scores, we can train the AffRankNet+ model by minimizing (\ref{eq:empirical_error_lupi}) with respect to the model parameters (NN weights). After training, the AffRankNet+ model computes a preference function $\hat{h}^*(x)$ that outputs the preference score for each data point $x$.

\begin{figure}[!tb]
	\begin{minipage}{1.0\linewidth}
		\centering
		\centerline{\includegraphics[trim=0 80 0 0 ,width=1.0\linewidth]{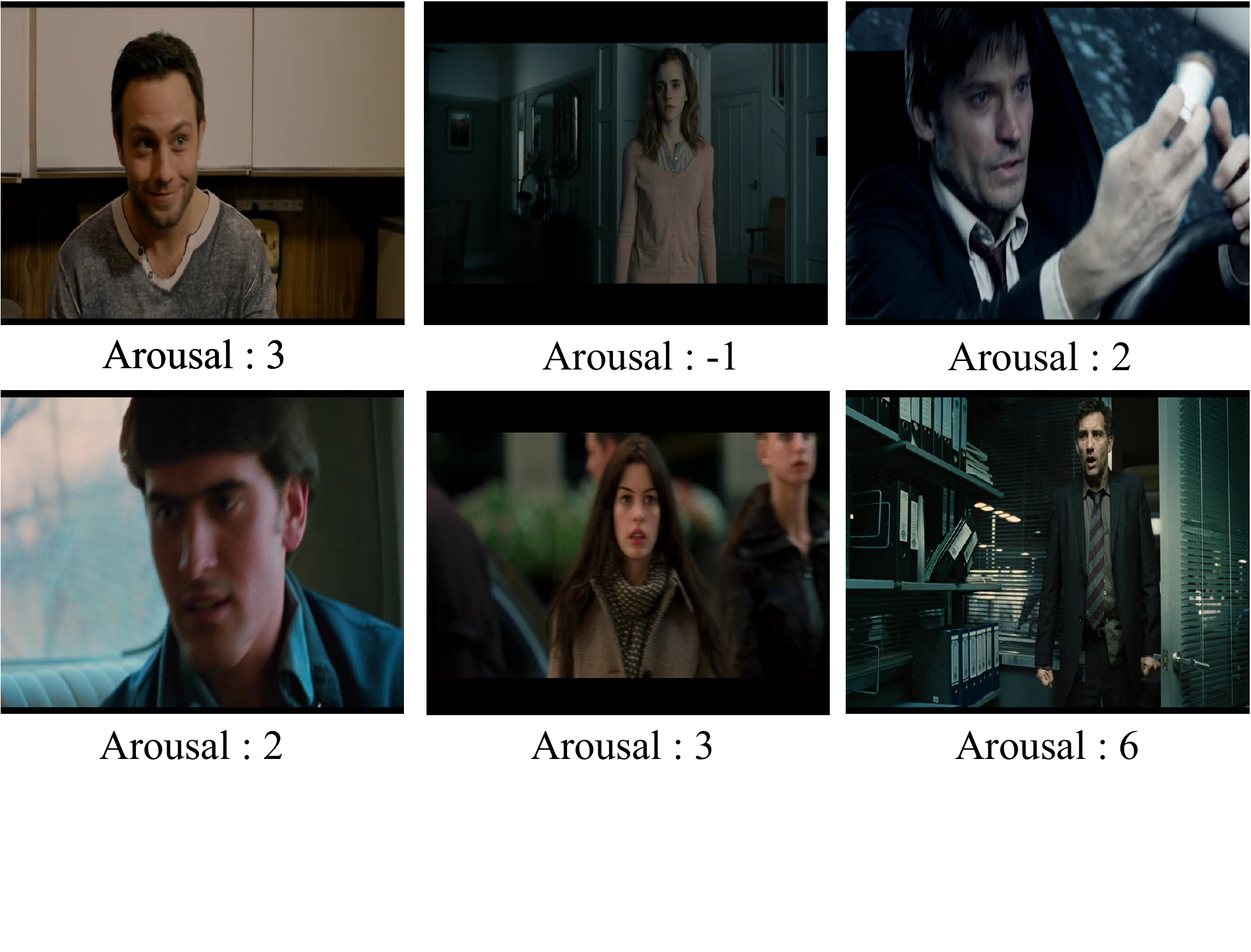}}
	\end{minipage}
	\caption{Indicative frames from the Afew-VA dataset along with their annotation for the arousal dimension.}
	\label{fig:afewva}
\end{figure}

\section{Experimental Setting and Evaluation}
In this section, we present the employed dataset including the training and test sets construction, the architecture of the AffRankNet+ model and the training details, and finally, the performance evaluation results. 

\begin{figure*}[!tb]
	\begin{minipage}{0.98\linewidth}
		\centering
		\centerline{\includegraphics[trim=70 120 50 10 ,width=0.95\linewidth]{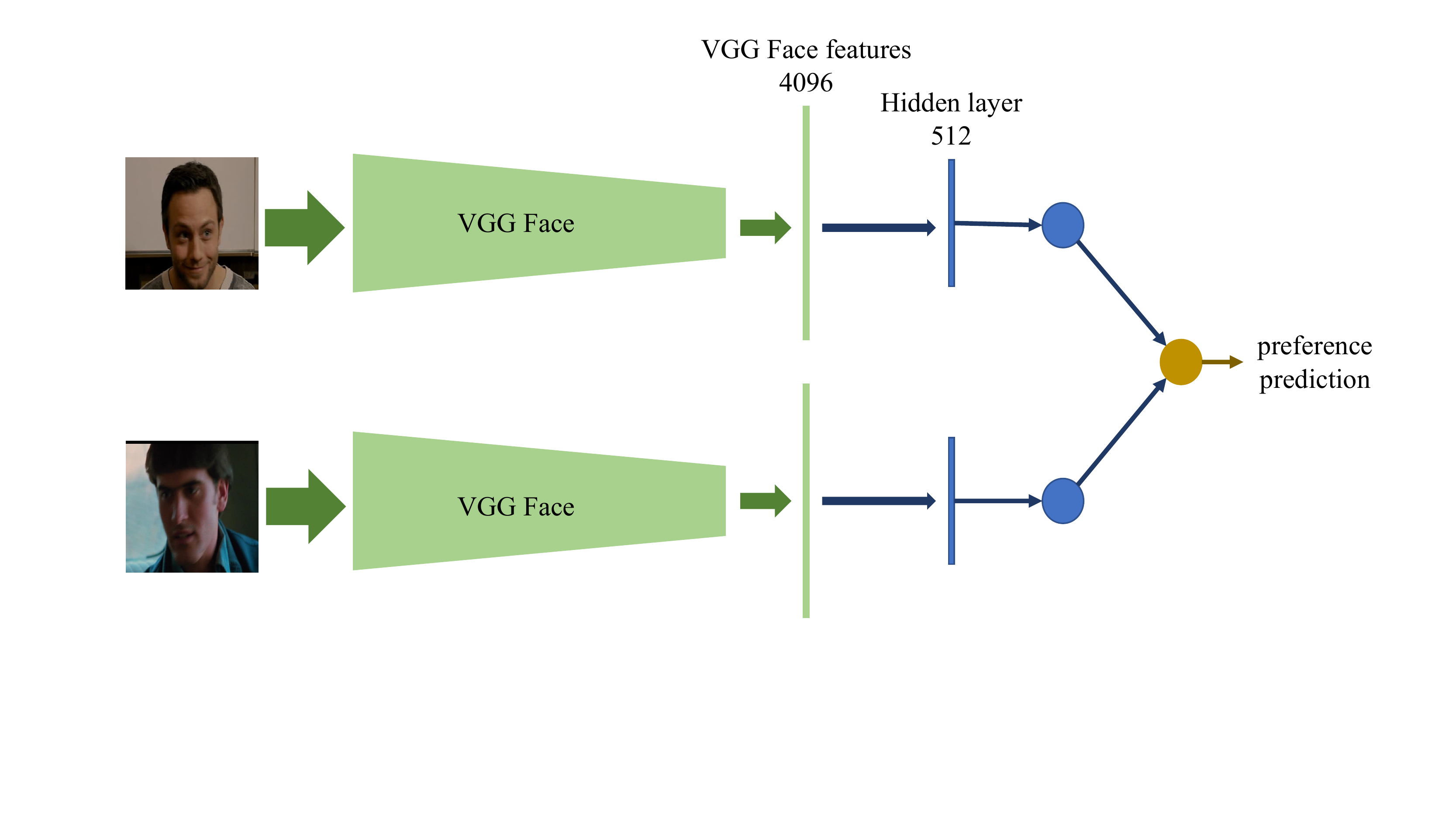}}
	\end{minipage}
	\caption{The architecture of AffRankNet+ model.}
	\label{fig:architecture}
\end{figure*}

\subsection{Dataset}
\label{ssec:dataset}
For evaluating the proposed AffRankNet+ model, we use the Afew-VA public available dataset \cite{kossaifi2017afew}. That dataset consists of 600 videos from films that range from 10 to 120 frames. The collected videos display various facial expressions. Each of the videos is annotated per frame, in terms of valence and arousal level, in the integer range [-10, 10].  In this study, which serves as a proof of concept, we consider only the arousal annotations. Therefore our objective is to rank the video frames based on the arousal level. Fig. \ref{fig:afewva} presents six indicative frames from the employed dataset along with their arousal annotation. We should note that the target variables for the Afew-VA dataset are \textit{subjectively} defined and thus their estimation is better suited within a ranking setting \cite{yannakakis2018ordinal, martinez2014don}.

The Afew-VA dataset, along with the frames, provides 68 facial landmark points. We use those landmark points to detect and crop the face. After cropping the face, we create a vector representation of the facial images using the features produced by the VGG-Face neural network \cite{parkhi2015deep} pre-trained for face recognition. Moreover, since the integer arousal annotation can be seen as preference scores, in our case the function $g(\cdot)$ in loss (\ref{eq:empirical_error_lupi}) is the identity function.

After defining the vector representation of frames and the form of privileged information, we have to construct the training and test sets in the form of (\ref{eq:sample_lupi}) for training and evaluating the performance of our model. To do so, in the first place, we split the Afew-VA dataset into two sets following the \textit{group} holdout scheme. This way, we can be sure that frames corresponding to the same video will be present either in the training set or the testing set, but not in both. Then, we compare the arousal annotation values of all pairs of points that belong to the same set and include in the training (test) set the pairs whose annotation difference is larger than a threshold. That threshold can be seen as a preference uncertainty bound which avoids resulting in a ranking model that its output is affected by trivial input differences. Such a threshold is commonly used when ranking algorithms are used for affect modelling (see for example \cite{makantasis2021pixels}). In this study, we set the value for that threshold equal to 4 following a trial-and-error procedure. The value of the threshold above balances, on the one hand, the richness of the data, and on the other, the size of the dataset, which highly affects the computational cost for training the model. 

\subsection{Architecture of AffRankNet+ and Training Details}
\label{ssec:details}

As mentioned before, the AffRankNet+ model is a two-stream neural network. The two streams have exactly the same topology and tied weights. In this study, the AffRankNet+ model uses the pre-trained VGG Face as a backbone network for constructing features from the face images. The VGG Face network builds features with 4096 elements, which then are fed to a fully connected feedforward neural network with one hidden layer of 512 neurons. We keep fixed the weights of the VGG Face feature construction network during training and modify only the weights of the subsequent fully connected feedforward neural network. Fig. \ref{fig:architecture} visually presents the architecture of the proposed AffRankNet+ model. The green part corresponds to the VGG Face backbone network, while the blue to the trainable part of the AffRankNet+.

We conduct experiments with varying training sizes; that is, 5\%, 10\%, and 20\% of the whole dataset are used for training and the rest for testing. 10\% of the training set is used as the validation set to activate early stopping criteria; the training stops after 15 epochs without validation loss improvement. We choose to use a small percentage of the whole dataset for training since the exploitation of privileged information reduces the sample complexity of learning, enabling the efficient training of the model using a small number of labelled data. For each training set size, we run ten experiments following the group holdout cross-validation scheme (see Section \ref{ssec:dataset}). For all the experiments, we keep $\tau$ parameter fixed equal to 1 (further investigations regarding the effect of $\tau$ on the model's performance are left for our future work). Finally, for updating the model's weights, we use the Adam optimizer with a learning rate equal to 0.001.

\subsection{Evaluation Results}
Following the experimental setting described above, we evaluate the ranking performance of the proposed model. The evaluation takes place in terms of average Pearson's correlation coefficient ($r$) and average Kendall's tau ($\uptau$) since these metrics are widely used for evaluating ranking algorithms \cite{kossaifi2017afew, melhart2020study}. 

First, we investigate the effect of $\lambda$ parameter (see equation (\ref{eq:empirical_error_lupi})) on the performance of the model by running experiments for different values of $\lambda$, namely $\lambda=0.3$, $0,5$, $0.8$. Second, we compare the performance of the AffRankNet+ model against the performance of the RankNet model, which does not exploit privileged information. To conduct a fair comparison, the two models have the same architecture and are trained and evaluated on the same sets of data points.

Tables \ref{tab:pearson} and \ref{tab:kendall} present the performance of AffRankNet+ model in terms of Pearson's correlation coefficient ($r$) and Kendall's tau ($\uptau$), respectively. We can see that for small-sized datasets, larger values of $\lambda$ parameter yield better model's performance both in terms of Pearson's $r$ and Kendall's $\uptau$. The obtained results agree with the formulation of loss in equation (\ref{eq:empirical_error_lupi}) used by the AffRankNet+ model. The number of points in a dataset and the uncertainty about the preference scores are inversely proportional. As mentioned before, parameter $\lambda$ quantifies the degree to which the model should trust the privileged information from the preference scores. Therefore, when the uncertainty is large, the model achieves better results by weighting more the first (preference relations) term in equation (\ref{eq:empirical_error_lupi}). On the contrary, when the size of the dataset is adequately large, the second term in (\ref{eq:empirical_error_lupi}), associated with preference scores, is more important and smaller values for parameter $\lambda$ yield better ranking results. 

\begin{table}[t]
	\caption{AffRankNet+ performance in terms of Pearson's correlation coefficient ($r$) using three different values for parameter $\lambda$.}
	\begin{tabular}{|r||c|c|c|} \hline
		& \begin{tabular}[c]{@{}c@{}}Dataset size\\ 5\%\end{tabular} & \begin{tabular}[c]{@{}c@{}}Dataset Size \\ 10\%\end{tabular} & \begin{tabular}[c]{@{}c@{}}Dataset size\\ 20\%\end{tabular} \\ \hline
		AffRankNet+ ($\lambda=0.3$) & 0.262   & 0.302   & \textbf{0.293} \\ \hline
		AffRankNet+ ($\lambda=0.5$) & 0.258   & 0.312   & 0.289 \\ \hline
		AffRankNet+ ($\lambda=0.8$) & \textbf{0.263}   & \textbf{0.322}   & 0.284 \\ \hline   
	\end{tabular}
	\label{tab:pearson}
\end{table}

\begin{table}[t]
	\caption{AffRankNet+ performance in terms of Kendall's tau coefficient ($\uptau$) using three different values for parameter $\lambda$.}
	\begin{tabular}{|r||c|c|c|} \hline
		& \begin{tabular}[c]{@{}c@{}}Dataset size\\ 5\%\end{tabular} & \begin{tabular}[c]{@{}c@{}}Dataset Size \\ 10\%\end{tabular} & \begin{tabular}[c]{@{}c@{}}Dataset size\\ 20\%\end{tabular} \\ \hline
		AffRankNet+ ($\lambda=0.3$) & 0.172   & 0.198   & \textbf{0.210} \\ \hline
		AffRankNet+ ($\lambda=0.5$) & 0.168   & 0.204   & 0.206 \\ \hline
		AffRankNet+ ($\lambda=0.8$) & \textbf{0.179}   & \textbf{0.216}   & 0.191 \\ \hline    
	\end{tabular}
	\label{tab:kendall}
\end{table}

In the next set of experiments, we compare the performance of the proposed AffRankNet+ model against the RankNet model, which does not exploit privilege information. As mentioned above, the two compared models have exactly the same architecture and are trained/validated on exactly the same data points. This way, any difference in the performance of the two models would be due to the exploitation of privileged information that comes from the preference scores. We should mention that for the AffRankNet+ model, we use the best value for the $\lambda$ parameter based on our previous experiment, that is $\lambda=0.8$ for dataset sizes 5\% and 10\% and $\lambda=0.3$ for 20\% dataset size.

\begin{figure}[!tb]
	\begin{minipage}{1.0\linewidth}
		\centering
		\centerline{\includegraphics[width=1.0\linewidth]{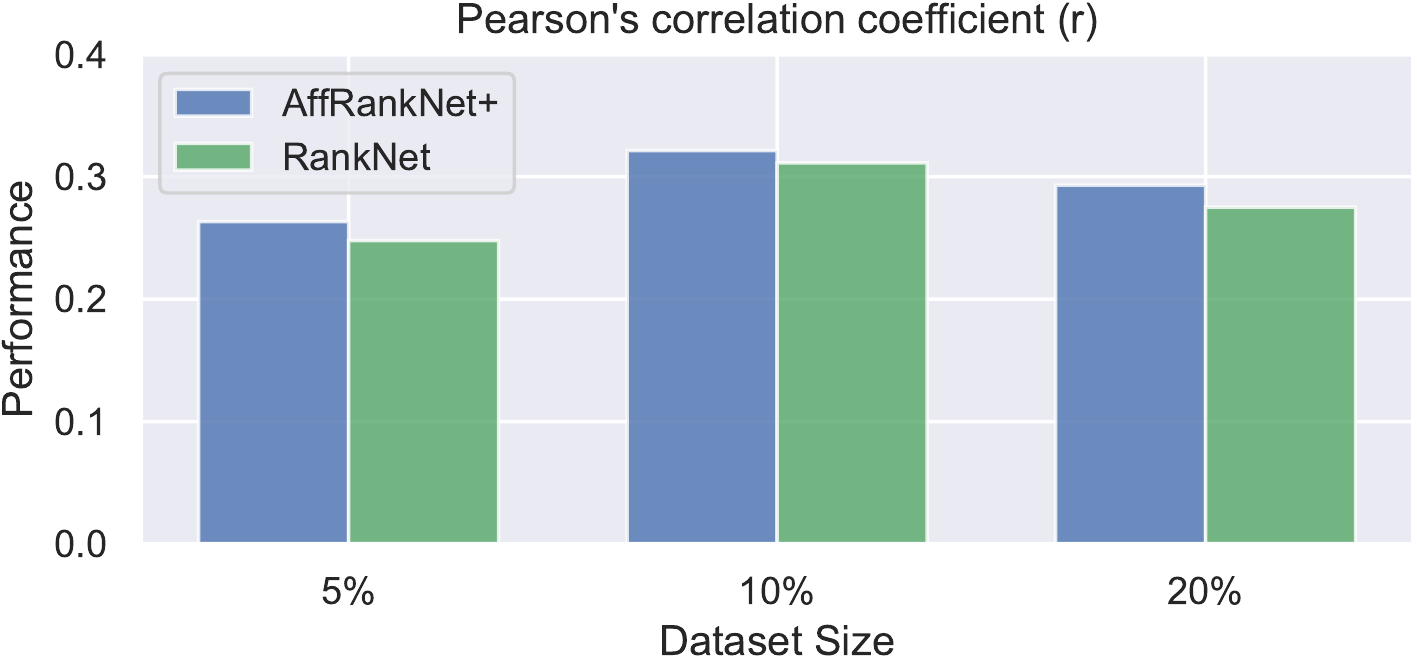}}
	\end{minipage} 
	
	\vspace{0.1in}
	
	\begin{minipage}{1.0\linewidth}
		\centering
		\centerline{\includegraphics[width=1.0\linewidth]{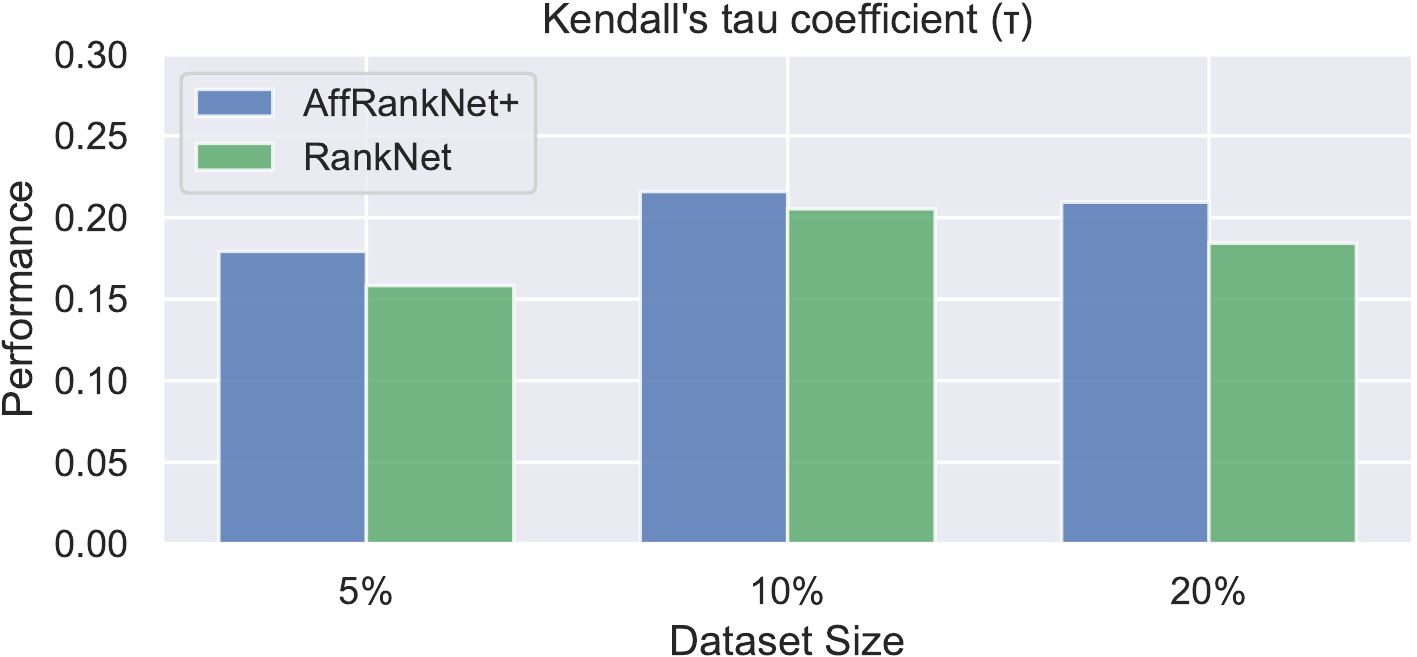}}
	\end{minipage}
	\caption{The architecture of AffRankNet+ model.}
	\label{fig:performance}
\end{figure}

Fig.\ref{fig:performance} presents the results from the comparison above. No matter the size of the dataset, the AffRankNet+ model achieves better performance in terms of both metrics than the RankNet model. Moreover, we test whether the improvement in performance achieved by using privileged information is statistically significant or not. Since we run ten experiments for each dataset size following the group holdout cross-validation scheme, we collect each fold's models' performance. Then, we test the null hypothesis the performances of the two models come from the same distribution. To do so, we conduct paired t-tests. Based on the t-tests outcomes, for all dataset sizes, we can reject the null hypothesis at a significance level of 0.05. Therefore, we can safely conclude that the exploitation of privileged information can significantly boost the performance of a ranking model.  

At this point, we should stress out that this study is not focusing on proposing a state-of-the-art ranking model for that specific dataset. Instead, it focuses on the importance of exploiting privileged information and on presenting a general methodology for ranking problems. Therefore the Afew-VA dataset is used as a proof-of-concept.

\section{Conclusions}
This study introduces the AffRankNet+ model for ranking affect states using privileged information associated with preference scores. To the best of our knowledge, this is the first time that a ranking model based on neural networks follows the LUPI paradigm. Although this study considers that preference scores are available, the formulation of the AffRankNet+ model is general to allow other types of information associated with preference scores to be used as privileged. For example, facial action units can be used as privileged information for learning to rank using the pixels' information solely from images of faces. We tested the ranking performance of AffRankNet+ on the public available Afew-VA dataset and compared it against the RankNet model. To conduct a fair comparison, both the AffRankNet+ and RankNet models have the same architecture and are trained/validated on the same data points. The experimental results emphasize the importance of privileged information by indicating that AffRankNet+, when appropriately parameterized, can perform significantly better than the RankNet model.


\end{document}